\documentclass[conference]{IEEEtran}
\IEEEoverridecommandlockouts
\usepackage{cite}
\usepackage{amsmath,amssymb,amsfonts}
\usepackage{algorithmicx}
\usepackage{graphicx}
\usepackage{textcomp}
\usepackage{xcolor}
\usepackage{multirow}
\usepackage{cases}
\usepackage{algorithm}
\usepackage{algpseudocode}
\def\BibTeX{{\rm B\kern-.05em{\sc i\kern-.025em b}\kern-.08em
    T\kern-.1667em\lower.7ex\hbox{E}\kern-.125emX}}
\begin{document}

\title{A Generalized Zero-Shot Framework for Emotion Recognition from Body Gestures}


\author{\IEEEauthorblockN{Jinting Wu\IEEEauthorrefmark{1}\IEEEauthorrefmark{2}, Yujia Zhang\IEEEauthorrefmark{1}, Xiaoguang Zhao\IEEEauthorrefmark{1}, Wenbin Gao\IEEEauthorrefmark{3}}
\IEEEauthorblockA{\IEEEauthorrefmark{1}\textit{The State Key Laboratory of Management and Control for Complex Systems} \\
\textit{Institute of Automation, Chinese Academy of Sciences}\\
\IEEEauthorrefmark{2}\textit{School of Artificial Intelligence}\\
\textit{University of Chinese Academy of Sciences}\\
\IEEEauthorrefmark{3}\textit{Key Laboratory of Mental Health}\\
\textit{Institute of Psychology, Chinese Academy of Sciences}\\
Beijing, China \\
Email: wujinting2016@ia.ac.cn, zhangyujia2014@ia.ac.cn, xiaoguang.zhao@ia.ac.cn, gaowb@psych.ac.cn}
}

\maketitle

\begin{abstract}
Although automatic emotion recognition from facial expressions and speech has made remarkable progress, emotion recognition from body gestures has not been thoroughly explored. People often use a variety of body language to express emotions, and it is difficult to enumerate all emotional body gestures and collect enough samples for each category. Therefore, recognizing new emotional body gestures is critical for better understanding human emotions. However, the existing methods fail to accurately determine which emotional state a new body gesture belongs to. In order to solve this problem, we introduce a Generalized Zero-Shot Learning (GZSL) framework, which consists of three branches to infer the emotional state of the new body gestures with only their semantic descriptions. The first branch is a Prototype-Based Detector (PBD) which is used to determine whether an sample belongs to a seen body gesture category and obtain the prediction results of the samples from the seen categories. The second branch is a Stacked AutoEncoder (StAE) with manifold regularization, which utilizes semantic representations to predict samples from unseen categories. Note that both of the above branches are for body gesture recognition. We further add an emotion classifier with a softmax layer as the third branch in order to better learn the feature representations for this emotion classification task. The input features for these three branches are learned by a shared feature extraction network, i.e., a Bidirectional Long Short-Term Memory Networks (BLSTM) with a self-attention module. We treat these three branches as subtasks and use multi-task learning strategies for joint training. The comprehensive experiments are conducted on an emotion recognition dataset based on body gestures, and the performance of our framework is significantly superior to the traditional method of emotion classification and state-of-the-art zero-shot learning methods.
\end{abstract}


\section{Introduction}
In human-human interaction, body language is one of the most important emotional expressions. Knapp et al. proposed that body postures and movements are as important as facial expressions for the task of emotion analysis and understanding~\cite{em_theory}. At the same time, body gestures are also of great significance for the analysis of emotion intensity~\cite{em_theory2}. However, most existing works on emotion analysis are mainly based on facial expressions and speech. As an important factor to convey emotional information, body language has not been deeply studied in emotion recognition.

The research on emotion recognition from body language mainly focuses on how to design the artificial features of body motion that are most conducive to emotion recognition. In recent years, with the development of deep learning techniques, many researchers have developed deep networks to exploit the most effective representations of body motion and achieved more accurate emotion classification results. These algorithms mainly apply the existing body detection and feature extraction techniques to the emotion classification task, but the relation between body gestures and emotions are not well established.

Another limiting factor of existing research is the lack of data due to difficulty in data collection. Most emotion recognition datasets based on body gestures contain only several hundreds of samples~\cite{Castellano2007,FABO,Kinect_dataset,MASR}, and most of the current research collects the actions performed by the experimenters in a laboratory environment. Body gestures in this collecting method mostly need to be specified by the experiment designer in advance, and the number of the gesture categories is small. However, there are rich meanings in body gestures, and the ways people express their emotions are different. When a new body gesture appears during the test, the recognition algorithm will easily make mistakes. More specifically, existing algorithms cannot give correct prediction results when an new emotional state appears. One method to address the above problem is to expand the training dataset to include as many emotional body gestures as possible. However, it is costly and laborious to collect the labeled data of all categories.

Zero-Shot Learning (ZSL) can establish associations between seen and unseen categories with side information such as attributes~\cite{firstZSL,SAE} and semantic vectors~\cite{CONSE}. It provides another solution to this problem, which is to recognize new body gesture categories using their semantic descriptions, and then infer the emotion categories from body gesture labels. Note that in ZSL tasks, training and test categories are strictly disjoint, and only data from the seen categories are used during training. However, when data from both seen and unseen categories are available during the test, ZSL methods have an inherent bias towards the seen categories. In other words, the ZSL classifiers tend to misidentify the samples from unseen categories as seen categories~\cite{action}. To solve this problem, GZSL, a more general task is proposed, where samples from the unseen categories are mixed with the seen categories during the test~\cite{review}. It aims to reduce the effect of such bias in a less restricted setting where training and test categories are not disjoint.

Although the ZSL approaches for object recognition~\cite{firstZSL,SAE,ESZSL} and action recognition~\cite{action,action2} have been largely investigated and achieved great success, there is no method applied in emotion recognition from body gestures under the ZSL or GZSL settings. Existing approaches are difficult to be directly applied to our task. Firstly, the recognition performance of these algorithms under the GZSL setting has not yet been satisfactory. How to effectively distinguish the seen and unseen categories and achieve high accuracy on both seen and unseen categories at the same time still needs to be studied. In addition, we propose to infer emotions by predicting categories of body gestures, but the existing action classifiers does not embed emotional information during training, which will cause large errors in emotion prediction.

Considering the above problems, this paper proposes a novel GZSL framework for emotion recognition from body gestures, which can recognize new emotional body gestures and new emotional states. The framework classifies the body gesture categories using their semantic descriptions, and then obtains emotion recognition results from body gesture labels. The GZSL algorithm in this framework is an extension of our previous work~\cite{mywork}, which is used for zero-shot hand gesture recognition. Based on the previous work, the algorithm in this paper uses a multi-head self-attention module in the feature extraction network to extract long-range temporal correlation information. Manifold regularization is also added to the zero-shot classifier, which is beneficial to construct the tight relations between geature space and semantic space. In addition, different from the existing GZSL algorithms, our framework adds an emotion classifier and jointly trains it with the above GZSL classifiers as subtasks, in order to learn features that are conducive to emotion classification during training. Adding attention modules to these subtasks and adjusting the learning rate of each subtask reseparately are also helpful to improve the accuracy of emoiton recognition.

The contributions of this paper are as follows:

\begin{itemize}
\item A novel framework with three branches is proposed for emotion recognition from body gestures. The first two branches are a Prototype-Based Detector (PBD) and a Stacked AutoEncoder (StAE), which are used to give the prediction results of the seen and unseen body gesture categories respectively. The third branch is an emotion classifier, which can enhance the generalization of the framework.

\item To the best of our knowledge, this is the first work attempting to recognize different emotions from body gestures under the GZSL setting, which is beneficial to infer a broader range of emotional states when new body gestures are given.

\item We adopt multi-task learning strategies, which take the three branches as subtasks and share low-level features for joint learning. Attention mechanism is also applied to these subtasks to weight features according to their importance.

\item We utilize an emotion dataset~\cite{MASR} which contains body movement data of multiple users, divide it into different body gesture categories, and design the semantic descriptions. Comprehensive experimental results on this dataset demonstrate the effectiveness of the proposed framework.
\end{itemize}

The rest of the paper is organized as follows. A brief review of emotion recognition from body gestures and zero-shot learning is given in Section~\ref{sec:Related Work}. The overall structure and main modules of our framework are presented in detail in Section~\ref{sec:Methodology}. Section~\ref{sec:Experiments} provides the experimental results and analysis. Finally, Section~\ref{sec:Conclusion} concludes the paper, and provides the discussion and the future work.

\section{Related Work}
\label{sec:Related Work}
\subsection{Emotion Recognition from Body Language}

The process of emotion recognition from body language generally includes human detection, pose estimation and tracking, feature extraction, and emotion classification. Human detection and pose estimation have been developed as research hotspots in computer vision area. In this paper, we mainly focus on the other two parts: feature extraction and emotion classification.

Most of the emotion recognition methods on body language focus on the whole body, upper body and gestures, where the whole body movements contain the most abundant information~\cite{emotion_survey}. In traditional feature extraction methods that focus on whole body movements, geometric models such as human skeleton joints are usually established to extract spatial features. Then, the motion features such as movement direction and speed are also extracted for dynamic analysis. Castellano et al.~\cite{Castellano2007} extracted time-domain features by calculating parameters such as palm speed, acceleration, fluidity of movement, and made use of the nearest neighbor algorithm, decision tree and Bayesian network for classification. Behoora et al.~\cite{team} used RGB-D cameras to obtain users' skeletal data, and then analyzed their emotions from body features such as joint position, speed, acceleration. The comparative classification algorithms included decision tree, IBK classifier, random forest and naive Bayes algorithm. Piana et al.~\cite{Adaptive} used Qualisys acquisition system and Microsoft Kinect to collect three-dimensional motion data of the whole body, and automatically extracted global expression features at different levels from joint movements. Then, an abstraction layer based on dictionary learning was used to further process these motion features, and finally SVM was utilized for real-time classification.

In recent years, deep learning has also been gradually applied to the research of emotion recognition from body gestures. Barros et al.~\cite{Barros2015} proposed a Multichannel Convolutional Neural Network (MCCNN), which took grayscale images and images processed by two different Sobel filters as input to a three-channel convolutional network to extract spatio-temporal features. Multi-channel features were integrated by a fully connected layer, and the emotional states were obtained through logistic regression. Ghayoumi et al.~\cite{Ghayoumi2016} used three CNNs to respectively model facial expressions, speech, and gestures in the application of robot interaction systems, and finally fused the features on the decision layer. Ly et al.~\cite{ly2018} proposed an end-to-end deep learning method. The hash algorithm was used to extract the key frames of the video, and then CNN and convolution LSTM network were used to extract the features. Sun et al.~\cite{Sun2019} processed the original videos to extract skeleton information and obtain the temporal segments of the videos. Then, CNN, BLSTM and PCA algorithms were combined for high-level spatio-temporal feature extraction and classification.

The above research can only recognize the emotion categories of the emotional body gestures which are seen during training. However, in our method, we propose a GZSL algorithm which can recognize unseen body gestures and classify the corresponding emotions.

\subsection{Zero-Shot Learning and Generalized Zero-Shot Learning}

The early works of zero-shot learning directly construct classifiers for attributes of seen and unseen classes. Lampert et al.~\cite{firstZSL} first proposed the task of zero-shot learning and introduced an attribute-based classification approach using high-level descriptions of all categories. Later, some other works propose to learn mappings from feature space to semantic space. For example, Norouzi et al.~\cite{CONSE} mapped images to class embeddings and estimated unseen labels by combining the embedding vectors of the most possible seen classes. Romera-Paredes et al.~\cite{ESZSL} developed a simple yet effective approach which learned the relationships between features, attributes and categories by adopting a two-layer linear model. More recently, Kodirov et al.~\cite{SAE} adopted an encoder-decoder paradigm that was learned with an additional reconstruction constraint to project a visual feature vector into the semantic space. Morgado and Vasconcelos~\cite{SCoRe} proposed two semantic constraints for recognition, namely loss-based regularizer and code-word regularizer, and achieved state-of-the-art performance. Some other methods further map both features and attributes to a shared space or align the semantic space and feature space. For example, Changpinyo et al.~\cite{SYNC} introduced ``phantom'' classes to align semantic space and the model space, and the classifiers for the phantom classes were used to synthesize classifiers for unseen classes via convex combinations.

The limitation of zero-shot learning is that all test data only come from unseen classes. Therefore, a generalized zero-shot learning setting is proposed where both seen and unseen classes are available during the test. Recently, many works have been proposed to address this task by alleviating the data imbalance between seen and unseen categories. For example, Xian et al.~\cite{f-CLSWGAN} proposed a generative adversarial network (GAN), which synthesizes CNN features of unseen classes based on class-level semantic information. Another GAN-based model achieved improvements in balancing accuracy between seen and unseen classes by combining visual-semantic mapping, semantic-visual mapping and metric learning~\cite{GDAN}. Some other approaches formulate the GZSL task as a cross-modal embedding problem. For instance, Felix et al.~\cite{multi} investigated a multi-modal based algorithm that integrated both visual and semantic Bayesian classifiers and promoted a balanced classification accuracy between seen and unseen classes. Schonfeld et al.~\cite{CADA-VAE} learned latent features of images and attributes via aligned Variational Autoencoders, and the features contained the essential multi-modal information which was associated with unseen classes. Although these methods mainly target the challenge of data imbalance between seen and unseen categories, the bias still exists due to the similar treatment of all categories. Therefore, some methods train detectors that can distinguish between seen and unseen categories. Bhattacharjee et al.~\cite{AEdetect} proposed a novel detector based on an autoencoder with a reconstruction loss and a triplet cosine embedding loss to determine whether an input sample belongs to a seen or unseen category. This detector greatly improved the recognition performance in novel categories. Mandal et al.~\cite{action} introduced two separate classifiers for seen and unseen action categories, and synthesized video features for unseen categories to train an out-of-distribution detector. Compared to these methods, our framework combines the classifier for seen classes and the detector as a branch, which has a simpler structure and is more convenient for training.

\section{Methodology}
\label{sec:Methodology}

\begin{figure*}
\centering
\includegraphics[width=17cm]{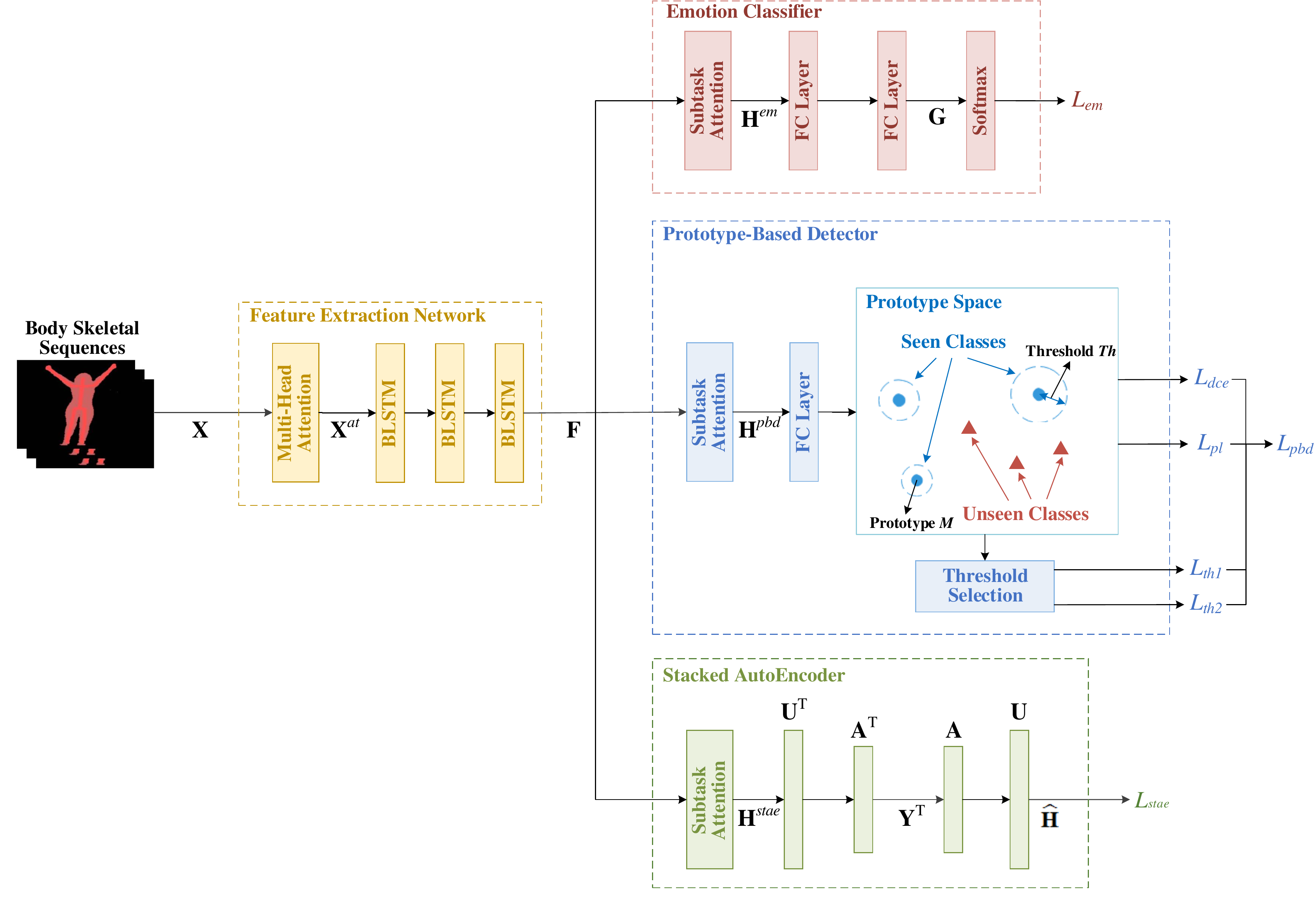}
\caption{An overview of the proposed framework. The framework consists of a shared feature extraction network and three branches: a Prototype-Based Detector (PBD), a Stacked AutoEncoder (StAE) and an emotion classifier.}
\label{fig1}
\end{figure*}

In this section, we first formalize the GZSL problem of emotion recognition, then introduce the details of our proposed framework. The proposed framework is shown in Fig.~\ref{fig1}, which consists of a shared feature extraction network and three branches: a Prototype-Based Detector (PBD), a Stacked AutoEncoder (StAE) and an emotion classifier. The feature extraction network is a Bidirectional Long Short-Term Memory Network (BLSTM) with a multi-head self-attention module, and the features are taken as the input to the three branches, respectively. The PBD trains prototypes to classify samples from the seen categories, and learns the threshold for each category to determine whether a sample belongs to this category. The StAE uses a two-layer AutoEncoder with manifold regularization to output predictions of samples from unseen classes. The emotion classifier with a softmax layer is used to enhance the generalization of the above two branches. Furthermore, multi-task learning strategies are applied to our framework and the loss functions of the three branches are jointly optimized.

\subsection{Problem Definition}

Let $D_s = \left\{ {{x_s},{y_s},{em(y_s)},{a_s(y_s)}}\right\}$ be the set of samples from seen body gesture classes, where ${x_s}$ is the body skeletal sequence, ${y_s}$ is the body gesture label of ${x_s}$ in the set of seen body gesture classes ${{\cal Y}^s}$, ${em(y_s)} \in \left\{1,\cdots ,C_{em} \right\}$ is the corresponding emotion label of ${y_s}$ where $C_{em}$ is the number of emotion categories, and ${a_s(y_s)}$ is the corresponding semantic representation of ${y_s}$. Similarly, the unseen data can be denoted as $D_u = \left\{ {{x_u},{y_u},{em(y_u)},{a_u(y_u)}}\right\}$, where the hand skeletal sequence ${x_u}$ is only available during testing, ${{y_u}}$ represents the body gesture label of ${x_u}$ in the set of unseen body gesture classes ${{\cal Y}^u}$, and ${{\cal Y}^u} \cap {{\cal Y}^s} = \emptyset $. The semantic representations of the seen and unseen classes for body gestures can be denoted as ${\bf{A}}_s=\left\{{a_s(i)}\right\}_{i=1}^{C_s}$ and ${\bf{A}}_u=\left\{{a_u(i)}\right\}_{i=1}^{C_u}$, where ${C_s}$ and ${C_u}$ are the number of seen and unseen body gesture categories, respectively.

In the GZSL task, all of the training samples come from the seen categories. The training data can be denoted as ${\bf{X}}_{tr}=\left\{{x_s^i}\right\}_{i=1}^{N_{tr}}$, where $N_{tr}$ is the number of training samples. In addition, samples in the test set are from both the seen and unseen categories, which can be denoted as ${\bf{X}}_{te}$. The goal of GZSL is to learn a classifier $f:{\cal X} \to {{\cal Y}^u} \cup {{\cal Y}^s}$. Note that in this emotion recognition task, the body gesture label ${\hat y}$ of the test sample $x$ is first obtained through the GZSL algorithm, its emotion recognition result $em(\hat y)$ is then achieved.

\subsection{Feature Extraction with Multi-Head Self-Attention Module}

The input sequences which include the position and orientation of the body's skeleton joints are captured by a Microsoft Kinect device. Multi-layer Bidirectional Long Short-Term Memory Networks (BLSTM)\cite{BLSTM} are used for feature extraction to simultaneously capture both the past and future contextual information. A BLSTM layer is composed of two LSTM layers (a forward one and a backward one). The outputs of the BLSTM are high-level features, which will be input into three branches.

In order to extract long-range temporal correlation information from the body gesture sequences, the multi-head self-attention module is added in front of the BLSTM. The multi-head attention based on the scaled dot-product attention was first proposed for machine translation tasks\cite{attention}. The multi-head structure is conducive to handling complex associations, and the interaction of distant frames can be achieved by the self-attention method.

The input of the scaled dot-product attention consists of three components: queries ${\bf{Q}} \in {{\mathbb{R}}^{{d_k} \times n}}$, keys ${\bf{K}} \in {{\mathbb{R}}^{{d_k} \times n}}$ and values ${\bf{V}} \in {{\mathbb{R}}^{{d_v} \times n}}$, where ${d_k}$, ${d_k}$ and ${d_v}$ are the dimensions of the queries, keys and values. In our framework, we set ${\bf{Q}}$, ${\bf{K}}$ and ${\bf{V}}$ using the same input skeleton sequence ${x} \in {{\mathbb{R}}^{{d_x} \times {l_x}}}$ to build a self-attention module, where ${d_x}$ and ${l_x}$ are the dimension and the length of ${x}$, respectively. The queries, keys and values are linearly projected $h$ times, and the outputs are concatenated and projected again to obtain the final attention result, which is given by:

\begin{equation}
{\rm{Attention}}\left( {{\bf{Q}},{\bf{K}},{\bf{V}}} \right) = {\rm{soft}}\max \left( {\frac{{{{\bf{Q}}^{\rm{T}}}{\bf{K}}}}{{\sqrt {{d_k}} }}} \right){{\bf{V}}^{\rm{T}}},
\end{equation}
\begin{equation}
{head_i} = {\rm{Attention}}\left( {{\bf{W}}_i^Q{x},{\bf{W}}_i^K{x},{\bf{W}}_i^V{x}} \right),
\end{equation}
\begin{equation}
\begin{split}
{{x}^{at}} &= {\rm{MultiHead}}\left( {{x},{x},{x}} \right) \\
&= {{\bf{W}}^{\bf{O}}}{\rm{Concat}}\left( {hea{d_1}, \ldots ,hea{d_h}} \right),
\end{split}
\end{equation}
where ${\bf{W}}_i^Q, {\bf{W}}_i^K, {\bf{W}}_i^V \in {{\mathbb{R}}^{{d_x} \times {d_x}}}$, and ${\bf{W}}_i^O \in {{\mathbb{R}}^{{d_x} \times h{d_x}}}$ are the projection matrices.

For the input data ${\bf{X}}=\left\{{{x}_i}\right\}_{i=1}^{N}$, the output of the attention module is denoted as ${\bf{X}}^{at}=\left\{{{x}^{at}_i}\right\}_{i=1}^{N}$. Then, the features extracted by the BLSTM network can be represented as:

\begin{equation}
{\bf{F}}{\rm{ = BLSTM}}\left( {{{\bf{X}}^{at}}} \right).
\label{eq:features}
\end{equation}

\subsection{Prototype-Based Detector (PBD) with Threshold Selection}

Traditionally, a softmax layer is added on top of the network for classification. However, the softmax-based approaches tend to misclassify unseen classes to seen classes. To solve this problem, Yang et al.\cite{CPL} proposed a convolutional prototype learning (CPL) framework, which can improve the robustness of classification. CPL is intended to learn a few prototypes using CNN features and predict classification labels by matching representations of test samples in the prototype space with the closest prototype. Inspired by this framework, we use a fully connected network to learn a prototype for each class and add an automatic threshold selection module to determine whether a test sequence belongs to a seen category.

The features input into the PBD model can be denoted as ${\bf{H}}^{pbd}$. The features are projected to the prototype space through FC layers. The projection of the $i^{th}$ sample in the prototype space is denoted as $p_i$, and the learned prototypes are defined as ${\bf{M}} = \left\{ {{m\left(k\right)}|k = 1, \cdots ,C_s} \right\}$, where ${m\left(k\right)}$ represents the prototype of the $k^{th}$ seen category. The parameters of FC layers and the prototypes ${\bf{M}}$ are jointly trained by minimizing the distance-based cross entropy (DCE) loss and prototype loss (PL).

The distance-based cross entropy (DCE) loss is based on the traditional cross entropy loss. It can be defined as:

\begin{equation}
{L_{dce}} =  - \frac{1}{{{N_{tr}}}}\sum\limits_{i = 1}^{{N_{tr}}} {\log \frac{{{e^{ - \gamma d\left( {p_i,m\left( {y_s^i} \right)} \right)}}}}{{\sum\nolimits_{k = 1}^{{C_s}} {{e^{ - \gamma d\left( {p_i,m\left( k \right)} \right)}}} }}},
\end{equation}
where $d\left( p_i,m\left( k \right) \right){\rm{ = }}\left\| p_i{\rm{ - }}m\left( k \right) \right\|_2^2$ computes the distance between the $p_i$ and the prototype of the $k^{th}$ category ${m\left( k \right)}$, ${y_s^i}$ is the label of the $i^{th}$ sample, and $\gamma $ is a hyper-parameter. Minimizing the DCE loss helps improve the classification accuracy and enhance the separability among different training categories.

The prototype loss (PL) is used as a normalization to regularize the model and enhance the intra-class compactness, which is defined as:

\begin{equation}
{L_{pl}} = \frac{1}{{{N_{tr}}}}\sum\limits_{i = 1}^{{N_{tr}}} {\left\| {p_i{\rm{ - }}m\left( {y_s^i} \right)} \right\|_2^2}.
\end{equation}

In the PBD branch, we also learn the distance thresholds for all prototypes. The sample whose distance from the closest prototype is greater than the corresponding threshold is classified as the unseen categories. The thresholds are defined as ${\bf{Th}} = \left\{ {{th\left(k\right)}|k = 1, \cdots ,C_s} \right\}$, where ${th\left(k\right)}$ represents the threshold corresponding to the prototype ${m\left(k\right)}$. The loss function of threshold selection is given by:

\begin{equation}
{L_{th1}} = \frac{1}{{{N_{tr}}}}\sum\limits_{i = 1}^{{N_{tr}}} {\left\{ {\begin{array}{*{20}{c}}
{0,\Delta {d_i} \le 0}\\
{\Delta {d_i},\Delta {d_i} > 0}
\end{array}} \right.},
\end{equation}
\begin{equation}
{L_{th2}} = \sum\limits_{k = 1}^{{C_s}} {{{\left( {th\left( k \right)} \right)}^2}},
\end{equation}
where $\Delta d_i = \left({\mathop {\min}\limits_k {d_i^k}}\right) - {th\left(\mathop {\arg\min}\limits_k {d_i^k}\right)}$, ${d_i^k}=d\left( {p_i,m\left( k \right)} \right)$, and ${th\left(\mathop {\arg\min}\limits_k {d_i^k}\right)}$ is the threshold of the category which the closest prototype belongs to. ${L_{th1}}$ tends to increase the thresholds to correctly classify more training samples, and ${L_{th2}}$ is used as the regularization to reduce the influence of outliers on threshold training.

In summary, the total loss function of the PBD module is defined as:

\begin{equation}
{L_{pbd}} = L_{dce} + {\beta _1}L_{pl} + {\beta _2}{L_{th1}} + {\beta _3}{L_{th2}},
\end{equation}
where ${\beta _1}$, $\beta_2$ and $\beta_3$ are hyper-parameters that weight the above loss functions.

\subsection{Stacked AutoEncoder (StAE)}

In GZSL tasks, semantic representations of both seen and unseen categories are available during the test. In order to recognize unseen gestures, a model needs to learn the relationship between the extracted features and the high-level semantic representations. In our previous work\cite{mywork}, we used a two-layer linear SAE to learn the semantic representation of each sample and match it with the semantic prototypes. However, the linear projection is difficult to model the complex correlation between the manifold structure of the visual feature space and the semantic space. In this section, we adopt a two-layer Stacked AutoEncoder (StAE) with manifold regularizations\cite{stae} to capture the manifold structures in both visual and semantic spaces and construct the tight relations of the two spaces.

During training, the input features ${\bf{H}}_{tr}^{stae}$ of the training samples are projected into the semantic space with the first layer of the encoder ${{\bf{U}}^{\rm{T}}} \in {{\mathbb{R}}^{{d_s} \times {d_h}}}$, where ${d_s}$ and ${d_h}$ are the dimension of the attributes and the input features. Then, the second layer uses the semantic representation matrix ${{\bf{A}}_s}^{\rm{T}} \in {{\mathbb{R}}^{{C_{s}} \times {d_s}}}$ of all seen semantic prototypes to establish the relationship between the semantic space and the label space. The structure of the decoder is symmetrical to the encoder, and we use the tied weights by setting the parameters of the decoder to ${{\bf{A}}_s}$ and ${\bf{U}}$. This auto-encoder aims to make the mapped embeddings in the label space close to the given labels, and at the same time, retain the original input information through the reconstruction of the decoder. The objective function is formulated as:

\begin{equation}
\mathop {\min }\limits_{\bf{U}} \left\| {{\bf{H}}_{tr}^{stae} - {\bf{U}}{{\bf{A}}_s}{\bf{Y}}_{tr}^{\rm{T}}} \right\|_F^2 + {\gamma _1}\left\| {{{\bf{A}}_s}^{\rm{T}}{{\bf{U}}^{\rm{T}}}{\bf{H}}_{tr}^{stae} - {\bf{Y}}_{tr}^{\rm{T}}} \right\|_F^2,
\end{equation}
where ${\bf{Y}}_{tr}$ is the one-hot vectors of the body gesture labels in the training set, and ${\gamma _1}$ is the hyper-parameter which weights the above two terms.

To model the manifold structure, we integrate two manifold regularizers\cite{stae} based on the instance graph and feature graph into this objective function. For instance graph ${G^I}$ with $n$ instances, the similarity matrix ${{\bf{W}}^I} \in {{\mathbb{R}}^{n \times n}}$ can be defined as:

\begin{small}
\begin{equation}
{\bf{W}}_{k,l}^I{\rm{ = }}\left\{ {\begin{array}{*{20}{l}}
{\cos \left( {{h_{*k}},{h_{*l}}} \right),{\rm{if~}}{h_{*k}} \in {N_q}\left( {{h_{*l}}} \right){\rm{~or~}}{h_{*l}} \in {N_q}\left( {{h_{*k}}} \right)}\\
{0,otherwise}
\end{array}} \right.,
\end{equation}
\end{small}where ${N_q}\left( {{h_{*k}}} \right)$ denotes the top-\emph{q} nearest neighbors of the ${k^{th}}$ instance ${h_{*k}}$. By denoting ${{\bf{Q}}^I} = diag\left( {\sum\nolimits_k {{\bf{W}}_{k,l}^I} } \right)$, the instance manifold regularizer can be defined as:

\begin{equation}
{R^I} = \frac{1}{2}\sum\limits_{k,l} {\left( {{{\left\| {{y_k} - {y_l}} \right\|}^2}} \right)} {\bf{W}}_{k,l}^I = tr\left( {{{\bf{Y}}^{\rm{T}}}\left( {{{\bf{Q}}^I} - {{\bf{W}}^I}} \right){\bf{Y}}} \right).
\end{equation}

Similarly, for feature graph ${G^F}$ with $d$ vertices which represent features, the similarity matrix ${{\bf{W}}^F} \in {{\mathbb{R}}^{d \times d}}$ can be defined as:

\begin{small}
\begin{equation}
{\bf{W}}_{k,l}^F{\rm{ = }}\left\{{\begin{array}{*{20}{l}}
{\cos \left( {{h_{k*}},{h_{l*}}} \right),{\rm{if~}}{h_{k*}} \in {N_r}\left( {{h_{l*}}} \right){\rm{~or~}}{h_{l*}} \in {N_r}\left( {{h_{k*}}} \right)}\\
{0,otherwise}
\end{array}} \right.,
\end{equation}
\end{small}where ${N_r}\left( {{h_{k*}}} \right)$ denotes the top-\emph{r} nearest neighbors of the ${k^{th}}$ dimension of feature ${h_{k*}}$. By denoting ${{\bf{Q}}^F} = diag\left( {\sum\nolimits_k {{\bf{W}}_{k,l}^F} } \right)$, the feature manifold regularizer can be defined as:

\begin{equation}
{R^F} = \frac{1}{2}\sum\limits_{k,l} {\left( {{{\left\| {{b_k} - {b_l}} \right\|}^2}} \right)} {\bf{W}}_{k,l}^F = tr\left( {{\bf{B}}\left( {{{\bf{Q}}^F} - {{\bf{W}}^F}} \right){\bf{B}}^{\rm{T}}} \right),
\end{equation}
where ${\bf{B}}{\rm{ = }}{{\bf{A}}^{\rm{T}}}{{\bf{U}}^{\rm{T}}}$. Then the final objective function of StAE can be formulated as:

\begin{equation}
\begin{split}
\mathop {\min }\limits_{\bf{U}} \left\| {{\bf{H}}_{tr}^{stae} - {\bf{U}}{{\bf{A}}_s}{\bf{Y}}_{tr}^{\rm{T}}} \right\|_F^2 + {\gamma _{\rm{1}}}\left\| {{{\bf{A}}_s}^{\rm{T}}{{\bf{U}}^{\rm{T}}}{\bf{H}}_{tr}^{stae} - {\bf{Y}}_{tr}^{\rm{T}}} \right\|_F^2 \\
{\rm{ + }}{\gamma _{\rm{2}}}{R_{tr}^I}{\rm{ + }}{\gamma _{\rm{3}}}{R_{tr}^F}.
\end{split}
\end{equation}

In the training stage, the instance manifold regularizer $R_{tr}^I$ for the training set is a constant variable. So we only need to optimize the following loss function:

\begin{small}
\begin{equation}
\begin{split}
{L_{stae}} = \left\| {{\bf{H}}_{tr}^{stae} - {\bf{U}}{{\bf{A}}_s}{\bf{Y}}_{tr}^{\rm{T}}} \right\|_F^2
+ {\gamma _{\rm{1}}}\left\| {{{\bf{A}}_s}^{\rm{T}}{{\bf{U}}^{\rm{T}}}{\bf{H}}_{tr}^{stae} - {\bf{Y}}_{tr}^{\rm{T}}} \right\|_F^2 \\
{\rm{ + }}{\gamma _{\rm{3}}}R_{tr}^F.
\end{split}
\end{equation}
\end{small}

During the test, because the StAE module is only used for the classification of unseen categories, we only use the semantic representations of the unseen categories ${\bf{A}}_u$. Considering that the feature manifold regularizer $R_{te}^F$ for the test set is a constant variable, a similar joint prediction mechanism based on test instances can be defined as:

\begin{equation}
\begin{split}
\mathop {\min }\limits_{{{\bf{Y}}_{te}}} \left\| {{\bf{H}}_{te}^{stae} - {\bf{U}}{{\bf{A}}_u}{\bf{Y}}_{te}^{\rm{T}}} \right\|_F^2 + {\gamma _{\rm{1}}}\left\| {{{\bf{A}}_u}^{\rm{T}}{{\bf{U}}^{\rm{T}}}{\bf{H}}_{te}^{stae} - {\bf{Y}}_{te}^{\rm{T}}} \right\|_F^2 \\
{\rm{ + }}{\gamma _{\rm{2}}}R_{te}^I,
\end{split}
\end{equation}
where ${\bf{H}}_{te}^{stae}$ is the features of test samples, $R_{te}^I$ represents instance manifold regularizer for the test set. By calculating derivative of the above equation and setting it to be zero, we can get the Sylvester equation as:

\begin{equation}
\begin{split}
\left( {{\bf{A}}_u}^{\rm{T}}{{\bf{U}}^{\rm{T}}}{\bf{U}}{{\bf{A}}_u} + {\gamma _{\rm{1}}}{{\bf{I}}^{C_u}} \right){\bf{Y}}_{te}^{\rm{T}} + {\gamma _{\rm{2}}}{\bf{Y}}_{te}^{\rm{T}} \left( {{\bf{Q}}_{te}^I - {\bf{W}}_{te}^I} \right) \\
= \left( {{\gamma _{\rm{1}}} + 1} \right){{\bf{A}}_u}^{\rm{T}}{{\bf{U}}^{\rm{T}}}{\bf{H}}_{te}^{stae}.
\end{split}
\end{equation}

Let ${\bf{{A}}}_{sy} = {{\bf{A}}_u}^{\rm{T}}{{\bf{U}}^{\rm{T}}}{\bf{U}}{{\bf{A}}_u} + {\gamma _{\rm{1}}}{{\bf{I}}^{C_u}}$, ${\bf{{B}}}_{sy} = {\gamma _{\rm{2}}} \left( {{\bf{Q}}_{te}^I - {\bf{W}}_{te}^I} \right)$ and ${\bf{{C}}}_{sy} = \left( {{\gamma _{\rm{1}}} + 1} \right){{\bf{A}}_u}^{\rm{T}}{{\bf{U}}^{\rm{T}}}{\bf{H}}_{te}^{stae}$. Then, this problem can be solved by using Bartels-Stewart algorithm \cite{bartels1972solution}:

\begin{equation}
{\bf{Y}}_{te}^{\rm{T}}{\rm{ = }}{\mathop{\rm sylvester}\nolimits} \left( {{{\bf{{A}}}_{sy}},{{\bf{B}}_{sy}},{{\bf{C}}_{sy}}} \right).
\end{equation}

The prediction result ${\bf{Y}}_{te}^{\rm{*}}$ is obtained by selecting the entity with the largest score of ${{\bf{Y}}_{te}}$:

\begin{equation}
\label{eq:stae_result}
{\bf{Y}}_{te}^{\rm{*}} = \arg \mathop {\max }\limits_l {\bf{Y}}_{te}^l,l \in [1,C],
\end{equation}
where ${\bf{Y}}_{te}^l$ is the value of the row corresponding to the ${l^{th}}$ category in ${{\bf{Y}}_{te}}$.

\subsection{Muti-task Learning for Emotion Recognition}

In the above sections, we introduce two branches: PBD and StAE, for the recognition of the seen and unseen body gesture categories, respectively. We can regard these two models as parallel subtasks to improve their generalization through joint learning. On the basis of this structure, multi-task learning strategies are introduced in our framework.

First, the subtasks share the feature extraction network, and jointly affect the parameters of the multi-head self-attention module and BLSTM during training. It is worth noting that the final task of our work is emotion classification. In order to learn the features which are more conducive to emotion recognition, we add an emotion classifier as the third subtask. This emotion recognition branch is not used for the result prediction in our framework, but only as an auxiliary task for the training of the other two branches. Specifically, it consists of multiple fully connected (FC) layers and a softmax layer. The cross-entropy loss function of this branch can be defined as:

\begin{equation}
{L_{em}} =  - \frac{1}{N_{tr}}\sum\limits_{i = 1}^{N_{tr}} {\log \left( {\frac{{{e^{g_{{em^i}}^i}}}}{{\sum\nolimits_{j = 1}^{{C_{em}}} {{e^{g_j^i}}} }}} \right)},
\end{equation}
where $g_j^i$ represents the component of the ${i^{th}}$ sample corresponding to the ${j^{th}}$ emotion category in the output vector of the FC layers, and similarly $g_{{em^i}}^i$ represents the component of the ${i^{th}}$ sample corresponding to its ground-truth emotion category.

Then, because the features extracted by BLSTM contribute differently to each subtask, the attention mechanism is applied to all branches to focus on the most important feature components for each subtask. It uses a hidden FC layer to calculate the distribution of attention and a softmax function for normalization, and finally multiplies the attention weights with the input to obtain new features. The attention function is defined as:

\begin{equation}
{\bf{H}} = \left( {{\rm{softmax}}\left( {{{\bf{W}}^{at}}{\bf{F}} + {{\bf{b}}^{at}}} \right)} \right){\bf{F}},
\end{equation}
where ${\bf{F}}$ is the features extracted by the BLSTM network, ${{\bf{W}}^{at}} \in {{\rm{R}}^{{d_f} \times {d_f}}}$ and ${{\bf{b}}^{at}} \in {{\rm{R}}^{{d_f} \times 1}}$ are parameters of the hidden FC layer, and ${d_f}$ is the dimension of the features ${\bf{F}}$. Three attention modules are trained for the above three branches, and the weighted features are represented as ${{\bf{H}}^{pbd}}$, ${{\bf{H}}^{stae}}$ and ${{\bf{H}}^{em}}$. Then they are used for branch training respectively.

However, it is difficult to train this network with a small amount of data. Especially, it is difficult to coordinate the loss weights and the learning rates of subtasks. Therefore, we divide the variables of our framework into four groups, which are shared networks and three branches. Then, we set different learning rates for the four groups, and adjust the weights of the loss function in each branch separately. The total loss function is defined as:

\begin{equation}
L{\rm{ = }}{L_{pbd}} + {\lambda _1}{L_{stae}} + {\lambda _2}{L_{em}},
\end{equation}
where ${\lambda _1}$ and ${\lambda _2}$ are hyper-parameters which weight the loss terms of the above three branches.

\subsection{Label Prediction}

During the test, the samples which belong to the seen body gesture categories are predicted by the PBD branch, and the prediction results of the samples from the unseen body gesture categories are given by the StAE branch. The process of the label prediction is shown in Alg.~\ref{alg1}, where $\Delta d$ is defined in (\ref{eq:d}).


\begin{algorithm}
  \caption{Emotion Recognition of Our Framework}
  \label{alg1}
  \begin{algorithmic}[1]
    \renewcommand{\algorithmicrequire}{\textbf{Input:}}
    \renewcommand{\algorithmicensure}{\textbf{Output:}}
    \Require
      test sequence $x$
    \Ensure
      emotion prediction $\hat E\left( x \right)$
    \State extract the body gesture features $\textbf{F}$
    \State obtain the projection $p$ of $x$ in the prototype space
    \State calculate the distance difference $\Delta d$
    \If{$\Delta d \le 0$}
      \State calculate the prediction of PBD branch $\varepsilon_{pbd}$
      \State $\hat E\left( x \right) \gets em \left( \varepsilon_{pbd} \right)$
    \Else
      \State calculate the prediction of StAE branch $\varepsilon_{stae}$
      \State $\hat E\left( x \right) \gets em \left( \varepsilon_{stae} \right)$
    \EndIf
    \State \Return{$\hat E\left( x \right)$}
  \end{algorithmic}
\end{algorithm}

For an input sample $x$, the extracted features are input into the PBD branch, and its projection $p$ in the prototype space is obtained. The body gesture category of the prototype which is closest to this projection is represented as:

\begin{equation}
\varepsilon_{pbd} \left( x \right) = \mathop {\arg \min }\limits_{k = 1}^{{C_s}} \left\| {{p}{\rm{ - }}m\left( k \right)} \right\|_2^2.
\end{equation}

Then, the model distinguishes the seen and unseen categories by comparing the minimum distance with the corresponding threshold:

\begin{equation}
\Delta d = \mathop {{\rm{min}}}\limits_{k = 1}^{{C_s}} \left\| {{p}{\rm{ - }}m\left( k \right)} \right\|_2^2 - th\left( {\varepsilon_{pbd} \left( x \right)} \right),
\label{eq:d}
\end{equation}
\begin{equation}
{\hat y}_{pbd}\left( x \right) = \left\{ {\begin{array}{*{20}{c}}
{\varepsilon_{pbd} \left( x \right),\Delta d  \le  0}\\
{{\rm{unseen}},\Delta d > 0},
\end{array}} \right.
\end{equation}
where ${\hat y}_{pbd}\left( x \right)$ represents the intermediate prediction result of PBD branch.

Furthermore, for the sample which is considered to be from an unseen category, its feature is further input into the StAE branch. The prediction result of the StAE branch $\varepsilon_{stae}\left( x \right)$ is given by (\ref{eq:stae_result}).

In summary, the emotion recognition result $\hat E\left( x \right)$ of the test sample $x$ is as follows:

\begin{equation}
\hat y\left( x \right) = \left\{ {\begin{array}{*{20}{c}}
{{\varepsilon _{pbd}}\left( x \right),\Delta d \le 0}\\
{{\varepsilon _{stae}}\left( x \right),\Delta d > 0},
\end{array}} \right.,
\end{equation}
\begin{equation}
\hat E\left( x \right)=em \left( {\hat y \left( x \right)} \right).
\end{equation}

\section{Experiments}
\label{sec:Experiments}
\subsection{Experiment Settings}
\subsubsection{Dataset}

We evaluate our framework on a public emotion dataset, MASR\cite{MASR}, which contains facial expressions and body gestures that are captured by Microsoft Kinect sensor in game scenes. In our experiments, we only use the gesture data to determine users' emotions. The skeleton data collected by Microsoft Kinect sensor is conducive to accurately extracting the 3D motion features of the human body, and it contains more information than the datasets that only collect the upper body data\cite{FABO,Kinect_dataset}. In this dataset, each subject performs two different body gestures of each emotion state in his own style, and each gesture was repeated 5 times. After removing incomplete and wrong samples, it contains 688 videos of 5 basic emotion states (anger, fear, happiness, sadness, surprise) which are performed by 15 subjects. Because our proposed framework recognizes emotions by classifying body gestures, we define the body gesture categories for each emotion state, and provide additional gesture labels to the dataset. There were 30 body gestures in our dataset. Due to the fact that different subjects have different habitual ways of expressing emotions, the amount of the samples in each gesture category is different. The body gesture categories in this paper and the number of their corresponding instances are shown in the Table \ref{table1}.

\begin{table}
\caption{Body Gesture Categories of the MASR Dataset}
\centering
\begin{tabular}{|c|c|c|c|c|}
\hline
\begin{tabular}[c]{@{}l@{}}Emotion\\ Category\end{tabular} & \begin{tabular}[c]{@{}c@{}}Body Gesture\\ Category\end{tabular} & \begin{tabular}[c]{@{}c@{}}The Instance\\ Number\end{tabular} & Partition1 & Partition2 \\ \hline
\multirow{6}{*}{Happy}                                     & 0                                                               & 67                                                            & seen   & seen   \\ \cline{2-5}
                                                           & 1                                                               & 4                                                             & unseen & seen   \\ \cline{2-5}
                                                           & 2                                                               & 5                                                             & seen   & seen   \\ \cline{2-5}
                                                           & 3                                                               & 15                                                            & unseen & seen   \\ \cline{2-5}
                                                           & 4                                                               & 34                                                            & seen   & seen   \\ \cline{2-5}
                                                           & 5                                                               & 20                                                            & seen   & seen   \\ \hline
\multirow{4}{*}{Sad}                                       & 6                                                               & 68                                                            & seen   & seen   \\ \cline{2-5}
                                                           & 7                                                               & 63                                                            & seen   & seen   \\ \cline{2-5}
                                                           & 8                                                               & 5                                                             & unseen & seen   \\ \cline{2-5}
                                                           & 9                                                               & 4                                                             & unseen & seen   \\ \hline
\multirow{7}{*}{Surprise}                                  & 10                                                              & 37                                                            & seen   & seen   \\ \cline{2-5}
                                                           & 11                                                              & 60                                                            & seen   & seen   \\ \cline{2-5}
                                                           & 12                                                              & 24                                                            & seen   & seen   \\ \cline{2-5}
                                                           & 13                                                              & 5                                                             & unseen & seen   \\ \cline{2-5}
                                                           & 14                                                              & 4                                                             & unseen & seen   \\ \cline{2-5}
                                                           & 15                                                              & 4                                                             & seen   & seen   \\ \cline{2-5}
                                                           & 16                                                              & 5                                                             & seen   & seen   \\ \hline
\multirow{7}{*}{Fear}                                      & 17                                                              & 37                                                            & seen   & seen   \\ \cline{2-5}
                                                           & 18                                                              & 25                                                            & seen   & seen   \\ \cline{2-5}
                                                           & 19                                                              & 15                                                            & unseen & seen   \\ \cline{2-5}
                                                           & 20                                                              & 28                                                            & seen   & seen   \\ \cline{2-5}
                                                           & 21                                                              & 5                                                             & unseen & seen   \\ \cline{2-5}
                                                           & 22                                                              & 10                                                            & seen   & seen   \\ \cline{2-5}
                                                           & 23                                                              & 7                                                             & seen   & seen   \\ \hline
\multirow{6}{*}{Anger}                                     & 24                                                              & 55                                                            & seen   & unseen \\ \cline{2-5}
                                                           & 25                                                              & 20                                                            & seen   & unseen \\ \cline{2-5}
                                                           & 26                                                              & 2                                                             & unseen & unseen \\ \cline{2-5}
                                                           & 27                                                              & 32                                                            & seen   & unseen \\ \cline{2-5}
                                                           & 28                                                              & 23                                                            & seen   & unseen \\ \cline{2-5}
                                                           & 29                                                              & 5                                                             & unseen & unseen \\ \hline
\end{tabular}
\label{table1}
\end{table}

\begin{figure}
\centering
\includegraphics[width=8.5cm]{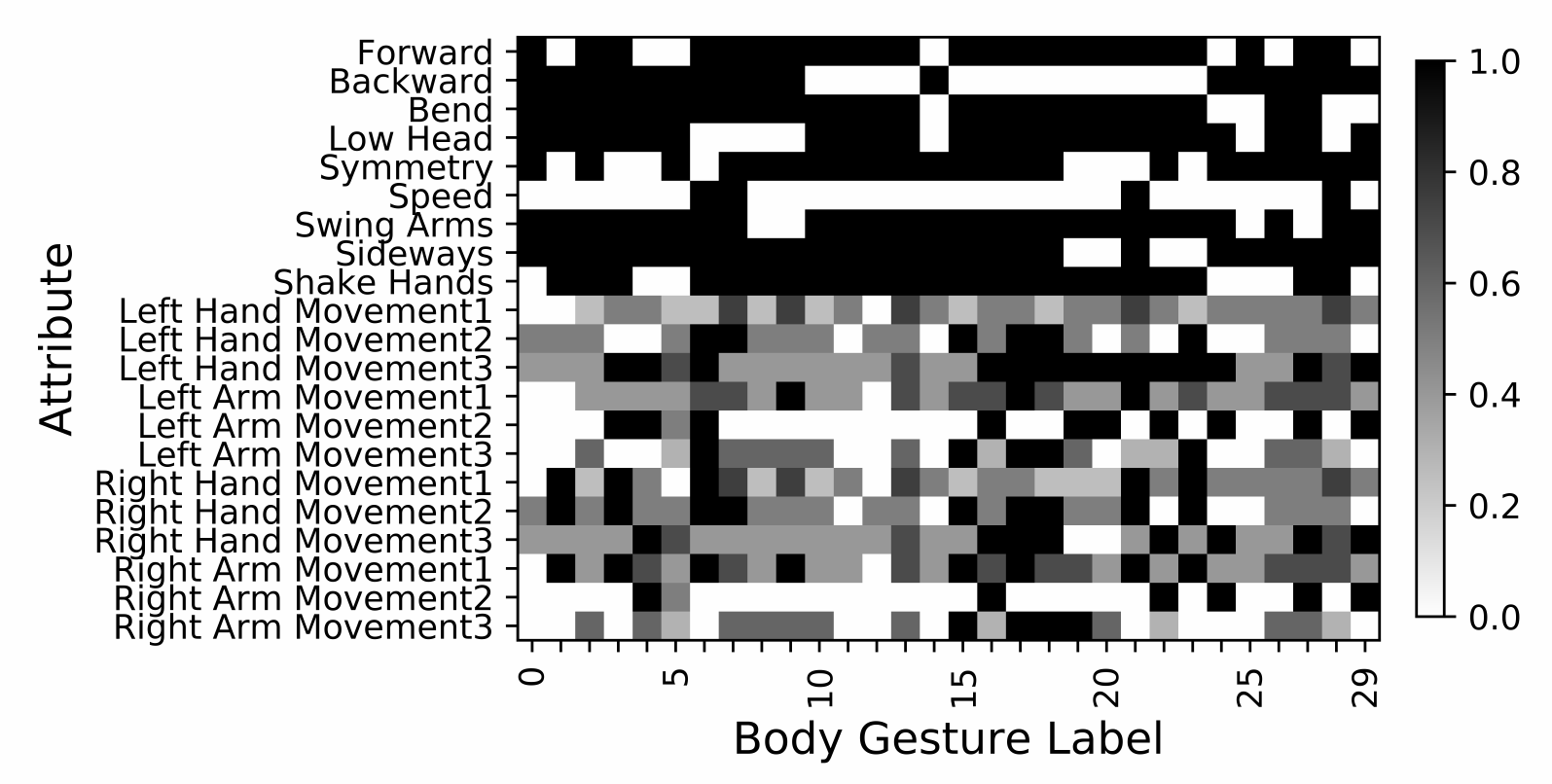}
\caption{Heat map of the categories and attributes.}
\label{fig3}
\end{figure}

Because the GZSL algorithms identify unseen categories by establishing the relationship between the semantic descriptions of seen and unseen body gestures, we design attribute vectors for all body gesture categories. Most of our attributes focus on the upper body, because upper body postures, especially arm movements, contain the most emotional information. The attributes which are used to encode the postures or movement trends of the hands and arms are continuous. The other attributes are binary, which are used to describe body movement trends, head movement, symmetry, speed, etc. The attributes are visualized in the form of heat map in Fig.~\ref{fig3}.

\subsubsection{Evaluation Metrics}

We adopt the top-1 accuracy to evaluate the models, and the top-1 accuracies of the seen gesture classes and unseen body gesture classes are denoted as $Ac{c_s}$ and $Ac{c_u}$. In order to ensure that both $Ac{c_s}$ and $Ac{c_u}$ are high enough, we also use harmonic mean $H$ for the performance comparison, which can be defined as:

\begin{equation}
H{\rm{ = }}\frac{{2 \times Ac{c_s} \times Ac{c_u}}}{{Ac{c_s}{\rm{ + }}Ac{c_u}}}.
\end{equation}

In addition to body gesture recognition, we also analyze the accuracy and harmonic mean in emotion recognition, which are denoted as $Acc_s^{em}$, $Acc_u^{em}$ and $H^{em}$.

\subsubsection{Dataset Partition}

In order to verify the performance of our framework, we propose two dataset partition settings.

In the first setting, we choose two body gestures as unseen categories in each emotion category. The seen and unseen categories are shown in the "Partition1" column in Table \ref{table1}. On the basis of ensuring that each category has at least one test sample, we randomly select approximately 1/10 samples from each seen category and add them into the test set. All samples of the unseen categories only exist in the test set. In total, the training set contains 554 samples, while the test set contains 70 samples from the seen categories and 64 samples from the unseen categories. The experimental results in this setting will be analyzed in Section \ref{sec:Partition1}.

The second setting is more specific, that is, the unseen gesture categories only come from the unseen emotion categories. In this setting, we regard all the body gestures of "angry" as unseen categories, as shown in the "Partition2" column in Table \ref{table1}. The training set consists of 493 samples, while the test set includes 58 samples from the seen categories and 137 samples from the unseen categories. In this setting, traditional emotion classifiers cannot give the correct results of unseen emotion categories. The experimental results in this setting are analyzed in Section \ref{sec:Partition2}.

\subsubsection{Implementation Details}

In feature extraction network, the number of the heads in the self-attention module is set to 5. We utilize a three-layer BLSTM network to extract features, and the numbers of forward and backward LSTM neurons are set to 64. The dimension of the output features of BLSTM is 128, and it remains unchanged after the attention module of each branch. In the branch of PBD, two FC layers which has 50 and 20 units respectively are used to project the features into the prototype space. The activation function we use in the network is ReLU, and the hyper-parameter $\gamma $ is set to 0.5. In StAE branch, the similarity matrices ${{\bf{W}}^I}$ and ${{\bf{W}}^F}$ are calculated based on the top-5 nearest neighbors. The emotion classifier consists of two FC layers, which has 50 and 5 units. During training, the batch size is set to 8. The Adam optimizer\cite{adam} is utilized to minimize the loss. Other hyper-parameters are experimentally set for different partitions of the dataset, which are shown in Table \ref{table2}.

\begin{table}
\caption{Hyper-parameters of Our Framework}
\centering
\begin{tabular}{c|c|c}
\hline
Parameters      & Values of Partition1 & Values of Partition2 \\ \hline
${lr_{shared}}$ & 0.0001               & 0.0001               \\
${lr_{pbd}}$    & 0.0001               & 0.00005              \\
${lr_{stae}}$   & 0.00002              & 0.00002              \\
${lr_{emotion}}$     & 0.00002              & 0.00005              \\
${\beta _1}$    & 4                    & 2                    \\
${\beta _2}$    & 0.1                  & 0.05                 \\
${\beta _3}$    & 1                    & 1                    \\
${\gamma _1}$   & 0.001                & 0.0001               \\
${\gamma _2}$   & 0.0001               & 0.0001               \\
${\gamma _3}$   & 0.1                  & 0.1                 \\
${\lambda _1}$  & 1                    & 1                    \\
${\lambda _2}$  & 4                    & 2                    \\ \hline
\end{tabular}
\label{table2}
\end{table}

\subsection{GZSL Experiments for Unseen Body Gesture Categories}
\label{sec:Partition1}

\subsubsection{State-of-the-art Comparisons}
In this section, we compare our proposed framework in the first data partition setting with the traditional emotion classifier and several state-of-the-art zero-shot learning methods.

\begin{table*}
\caption{Experimental Results of the State-of-the-art Comparisons in the First Data Partition Setting}
\centering
\begin{tabular}{l|c|c|c|c|c|c}
\hline
Methods            & $Acc_s$             & $Acc_u$             & $H$                    & $Acc_s^{em}$         & $Acc_u^{em}$         & $H^{em}$               \\ \hline
Emotion Classifier & /                & /                & /                    & \textbf{94.29\%} & 46.88\%          & 62.62\%              \\
CADA-VAE\cite{CADA-VAE} & 52.86\%          & 17.19\%          & 25.94\%              & 91.43\%          & 54.69\%          & 68.44\%           \\
f-CLSWGAN \cite{f-CLSWGAN} & 48.57\%          & 21.88\%          & 30.17\%              & 88.57\%          & 53.13\%          & 66.42\%           \\
Our Previous Work\cite{mywork} & 57.14\%          & 23.44\%          & 33.24\%              & 92.86\%          & 53.13\%          & 67.59\%           \\
Our Framework      & \textbf{61.43\%} & \textbf{31.25\%} & \textbf{41.43\%}    & 88.57\%          & \textbf{64.06\%} & \textbf{74.35\%} \\ \hline
\end{tabular}
\label{table3}
\end{table*}

The traditional emotion classifier is composed of a three-layer BLSTM and a softmax layer, and it only uses emotion labels for training. We also choose two state-of-the-art GZSL algorithms, CADA-VAE\cite{CADA-VAE} and f-CLSWGAN\cite{f-CLSWGAN}, for the comparisons on our dataset. The features of these algorithms are also obtained using a three-layer BLSTM. We further compare our framework with our previous work\cite{mywork} which includes a PBD and a Semantic Autoencoder (SAE). Experimental results are shown in Table \ref{table3}.

The traditional emotion classifier achieves the highest emotion recognition accuracy of the seen classes. However, the accuracy of the unseen classes is the lowest. Our framework outperforms the other zero-shot learning methods in body gesture recognition, and $Ac{c_s}$, $Ac{c_u}$ and $H$ are increased by 4.29\%, 7.81\% and 9.79\%, respectively. Because our main purpose is emotion classification, correctly predicting the emotion labels is much more important. Although our method has no advantage while testing the samples from the seen categories, $Acc_u^{em}$ increases from 54.69\% to 64.06\%, and $H^{em}$ increases from 68.44\% to 74.35\%. In our framework, two separate classifiers are used for gesture prediction, thus the influence of the inherent bias are reduced. This leads to a great improvement of our algorithm on the unseen categories. The application of the emotion branch and multi-task learning strategies enhances the adaptability of our framework for the task of emotion classification.

\subsubsection{Ablation Analysis}
\label{sec:ablation}

We analyze the different components in our framework to verify the effectiveness of these modules.

\begin{table*}
\caption{Experimental Results of Ablation Analysis in the First Data Partition Setting}
\centering
\begin{tabular}{l|c|c|c|c|c|c}
\hline
Methods                                & $Acc_s$          & $Acc_u$          & $H$              & $Acc_s^{em}$     & $Acc_u^{em}$     & $H^{em}$         \\ \hline
No Multi-head Self-attention Module    & 62.86\%          & 20.31\%          & 30.70\%          & 84.29\%          & 56.25\%          & 67.47\%          \\
No Attention Modules of Three Branches & \textbf{71.43\%} & 18.75\%          & 29.70\%          & 87.14\%          & 54.69\%          & 67.20\%          \\
No Emotion Branch                      & 70.00\%          & 26.56\%          & 38.51\%          & 84.29\%          & 57.81\%          & 68.58\%          \\
Same Learning Rate                     & 60.00\%          & 21.87\%          & 32.06\%          & 81.43\%          & 57.81\%          & 67.62\%          \\
Our Framework*                         & 61.43\%          & \textbf{31.25\%} & \textbf{41.43\%} & \textbf{88.57\%} & \textbf{64.06\%} & \textbf{74.35\%} \\ \hline
\end{tabular}
\label{table4}
\end{table*}

\textbf{Multi-head Self-attention Module.} We compare the framework without multi-head self-attention module to our framework in order to evaluate the effectiveness of this module. The experimental results are shown in Table \ref{table4} (line 1). We can observe that the accuracies of both gesture recognition and emotion recognition is lower after dropping the self-attention module in feature extraction. This  demonstrates that adding this module helps to extract more effective long-range features, and ultimately improves the classification performance.

\textbf{Attention Modules of Three Branches.} The experimental results without the branch attention modules are shown in Table \ref{table4} (line 2). We observe that the framework without attention modules has a higher accuracy for the seen gesture categories, but the performance on the unseen categories is lower. This demonstrates that the learned features are biased towards the PBD branch, and the training of StAE is restricted. The use of the attention mechanism helps to focus on the most important features for each task, so as to coordinate the needs and goals of all tasks and make each task achieve better results.

\textbf{Emotion Branch.} The accuracy of the framework without the emotion branch is shown in Table \ref{table4} (line 3). Although its $Acc_s$ is 8.57\% higher than our framework, the $Acc_s^{em}$ is 4.28\% lower. The results show that adding the emotion branch can improve the generalization ability of our framework and learn shared features that are more conducive to emotion classification.

\textbf{Learning Rate.} In this experiment, we set the same learning rate for all network parameters and optimize them uniformly. After testing, the optimal learning rate is 0.00005, and the corresponding results are shown in Table \ref{table4} (line 4). In fact, during training, the StAE branch converges faster and is prone to overfitting. Therefore, in our framework, the learning rates of the StAE branch are set to be smaller, and the learning rate of the PBD and shared network is set to be larger, so that all subtasks can achieve the best recognition results at the same training stage.

\subsubsection{Parameter Analysis}
In this section, we will focus on analyzing the weight coefficients of the loss functions of each branch to illustrate the role of each module on the overall framework and the impact of different weights on the performance of recognition.

\begin{figure*}
\centering
\includegraphics[width=17cm]{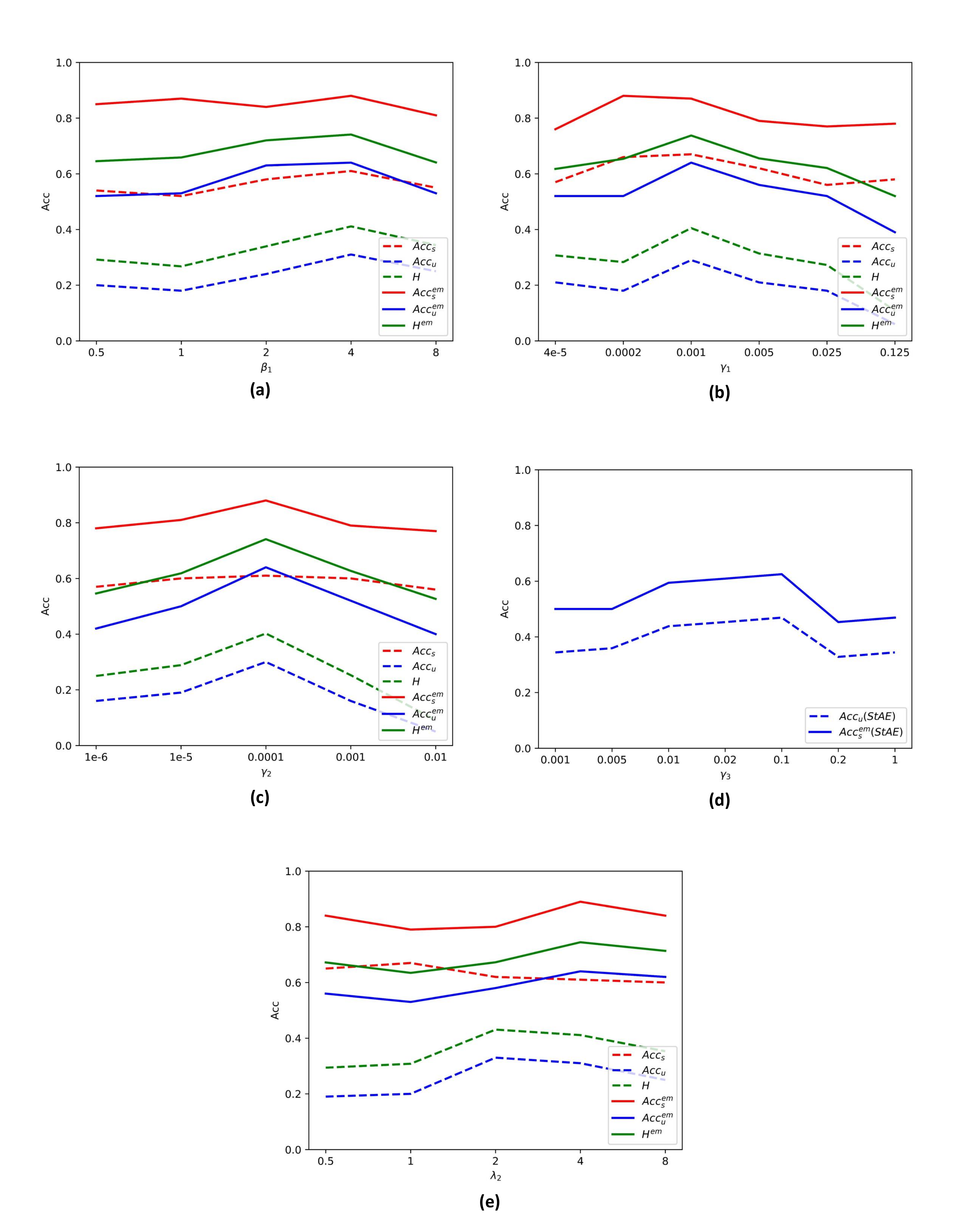}
\caption{Experimental results of parameter analysis. (a)Accuracies of different ${\beta _1}$; (b)Accuracies of different ${\gamma _1}$; (c)Accuracies of different ${\gamma _2}$; (d)Accuracies of different ${\gamma _3}$; (e)Accuracies of different ${\lambda _2}$.}
\label{fig4}
\end{figure*}

\textbf{Parameters of PBD.} In this section, we discuss the influence of the weights of the loss functions in PBD branch. First, ${\beta _1}$ is the weight of $loss_{pl}$, which will affects the intra-class compactness of the projections in the prototype space. As ${\beta _1}$ increases, the intra-class distance between the samples from the same categories decreases, and it is easier to distinguish seen and unseen categories using the thresholds. However, a too high ${\beta _1}$ will affect the optimization process of other loss functions, which in turn leads to a lower recognition accuracy of PBD and StAE. The results of different ${\beta _1}$ are shown in Fig.~\ref{fig4}(a). In addition, the learning of the thresholds determines the accuracy of distinguishing between the seen and unseen categories, which is of great significance for improving the recognition performance of the framework. The two parameters that are related to threshold learning are ${\beta _2}$ and ${\beta _3}$. The increase of ${\beta _2}$ tends to increase the thresholds to include as many training samples of the same category as possible, and the increase of ${\beta _3}$ reduces the thresholds by removing outliers. Because the thresholds are related to the compactness of the samples, these two parameters also need to be adjusted according to the value of ${\beta _1}$. In the experiment, we also find that too large ${\beta _2}$ and ${\beta _3}$ will cause the related loss functions to be dominant in the early stage of training, and make the algorithm difficult to converge. Therefore, we set ${\beta _2}$ and ${\beta _3}$ to 0 at the beginning of training, and then change the value of these two weights in the later stage of training.

\textbf{Parameters of StAE.} In StAE, we mainly analyze the influence of ${\gamma _1}$, ${\gamma _2}$, and ${\gamma _3}$ on the accuracy of the framework, and the experimental results are shown in Fig.~\ref{fig4}(b)-(d). The increase in ${\gamma _1}$ enhances the reconstruction ability of the Auto-encoder and better exploits the feature representations, which will result in an increase in the accuracy of StAE recognition. However, too large ${\gamma _1}$ will make the StAE branch unable to converge. When the value of ${\gamma _2}$ is very small, it is similar to the traditional Auto-encoder. With the increase of ${\gamma _2}$, the manifold structures in the visual space and the semantic space are better described. However, when ${\gamma _2}$ is too large, the algorithm is also difficult to converge. Finally, because ${\gamma _3}$ only appears in the test process, we give a more intuitive test result of the StAE branch without threshold discrimination. With the increase in ${\gamma _3}$, $Acc_u(StAE)$ and $Acc_u^{em}(StAE)$ have a synchronous trend of first increasing and then decreasing.

\textbf{Parameters of Emotion Classier.} In the Section \ref{sec:Partition1}, we compare our framework to the method without the emotion branch. The influence of different values of ${\lambda _2}$ is also shown in Fig.~\ref{fig4}(e). $Ac{c_s}$ decreases as ${\lambda _2}$ increases, because the emotion classier affects the convergence and the learned thresholds of PBD. However, as ${\lambda _2}$ increases, a stronger correlation between different body gestures of the same emotion category is established, and the generalization of the whole framework is enhanced. So the accuracies of emotion recognition have an upward trend in a certain range.

\subsection{GZSL Experiments for Unseen Emotion Categories}
\label{sec:Partition2}

\begin{table*}
\caption{The Experimental Results of the State-of-the-art Comparisons in the Second Data Partition Setting}
\centering
\begin{tabular}{l|c|c|c|c|c|c}
\hline
Methods            & $Acc_s$             & $Acc_u$             & $H$                    & $Acc_s^{em}$         & $Acc_u^{em}$         & $H^{em}$               \\ \hline
CADA-VAE\cite{CADA-VAE} & 62.07\%          & 20.44\%          & 30.75\%              & 72.41\%          & 55.47\%          & 62.82\%           \\
f-CLSWGAN \cite{f-CLSWGAN} & 60.34\%          & 22.63\%          & 32.92\%              & 68.97\%          & 67.15\%          & 68.05\%           \\
Our Previous Work\cite{mywork} & 63.79\%          & 22.63\%          & 33.41\%              & 70.69\%          & 68.61\%          & 69.63\%           \\
Our Framework      & \textbf{67.24\%} & \textbf{29.93\%} & \textbf{41.42\%}    & \textbf{82.76\%}          & \textbf{75.18\%} & \textbf{78.79\%} \\ \hline
\end{tabular}
\label{table5}
\end{table*}

In the second data partition setting, all of the test samples come from an unseen emotion category (anger). Because traditional emotion classifier cannot obtain the correct recognition results of unseen categories, we only compare several state-of-the-art zero-shot learning methods with our framework. The training parameters are shown in Table \ref{table2}, and experimental results are shown in Table \ref{table5}. Our algorithm achieves the best performance on all evaluation metrics. $Ac{c_s}$, $Ac{c_u}$ and $H$ are increased by 3.45\%, 7.3\% and 8.01\%, and the $Acc_s^{em}$, $Acc_u^{em}$ and $H^{em}$ are increased by 10.35\%, 6.57\% and 9.16\%, respectively. This demonstrates the effectiveness of our proposed method.

In fact, due to the small amount of samples in some training gesture categories, $Ac{c_u}$ and $Acc_s^{em}$ are lower than the experimental results in Section \ref{sec:Partition1}. The value of $Acc_u^{em}$ is higher since there is only one unseen emotion category in this experiment. As long as the test sample is detected as unseen, it can be classified into the correct emotion category.  As the number of unseen emotion categories increases, this value will decrease accordingly.

\section{Conclusion}
\label{sec:Conclusion}

In this paper, we introduce a novel generalized zero-shot learning (GZSL) framework for emotion recognition using body gestures. The framework contains three branches. The first branch is a Prototype-Based Detector (PBD), which is used to identify the seen gesture categories and determine whether the samples come from the unseen categories through the learned thresholds. The second branch is a Stacked AutoEncoder (StAE) with manifold regularization, which is used to classify the samples that are predicted to belong to the unseen categories. The third branch is an emotion classifier to enhance the generalization of the recognition framework. The experimental results on the MASR dataset verify that our algorithm is capable of handling unseen gesture categories or unseen emotion categories. In future work, we aim to investigate Few-Shot Learning (FSL) and Generative Adversarial Networks (GAN) to fully exploit the representative information in a small number of samples and further improve the performance. Besides, expressions, physiological signals and other modals could also be combined to make comprehensive judgments to reduce the interference of gesture ambiguity.

\section*{Acknowledgment}

This work is supported by the National Natural Science Foundation of China (Grant No. 61673378) and the National Key Research and Development Program of China (Grants No. 2019YFB1310601 and No. 2017YFC0820203).


%
%

\bibliographystyle{IEEEtran}
\bibliography{emotion}

\end{document}